\documentclass{article}
\usepackage{ijcai25}

\usepackage{times}
\usepackage{soul}
\usepackage{url}
\usepackage[hidelinks]{hyperref}
\usepackage[utf8]{inputenc}
\usepackage[small]{caption}
\usepackage{graphicx}
\usepackage{amsmath}
\usepackage{amsthm}
\usepackage{booktabs}
\usepackage{algorithm}
\usepackage{algorithmic}
\usepackage[switch]{lineno}
\usepackage{subcaption}

\title{Explainability Through Human-Centric Design for XAI in Lung Cancer Detection}

\author{
Amy Rafferty$^1$
\and
Rishi Ramaesh$^2$\and
Ajitha Rajan$^1$
\affiliations
$^1$University of Edinburgh, UK\\
$^2$NHS Lothian, UK\\
\emails
\ s1817812@ed.ac.uk
}

\begin{document}

\maketitle

\begin{abstract}
Deep learning models have shown promise in lung pathology detection from chest X-rays, but widespread clinical adoption remains limited due to opaque model decision-making. In prior work, we introduced ClinicXAI, a human-centric, expert-guided concept bottleneck model (CBM) designed for interpretable lung cancer diagnosis. We now extend that approach and present XpertXAI, a generalizable expert-driven model that preserves human-interpretable clinical concepts while scaling to detect multiple lung pathologies. Using a high-performing InceptionV3-based classifier and a public dataset of chest X-rays with radiology reports, we compare XpertXAI against leading post-hoc explainability methods and an unsupervised CBM, XCBs. We assess explanations through comparison with expert radiologist annotations and medical ground truth. Although XpertXAI is trained for multiple pathologies, our expert validation focuses on lung cancer. We find that existing techniques frequently fail to produce clinically meaningful explanations, omitting key diagnostic features and disagreeing with radiologist judgments. XpertXAI not only outperforms these baselines in predictive accuracy but also delivers concept-level explanations that better align with expert reasoning. While our focus remains on explainability in lung cancer detection, this work illustrates how human-centric model design can be effectively extended to broader diagnostic contexts — offering a scalable path toward clinically meaningful explainable AI in medical diagnostics.
\end{abstract}

\section{Introduction}

The black-box nature of modern deep learning (DL) models makes it challenging to trust and understand the rationale behind their decisions, particularly in high-stakes domains such as medical diagnostics \cite{blackboxbad}. Explainable artificial intelligence (XAI) aims to increase the transparency and trustworthiness of DL systems by offering interpretable insights into their decision-making processes. Broadly, XAI techniques fall into two categories: post-hoc methods, which are applied after model training, and ante-hoc methods, which are designed to make models intrinsically interpretable. Prior studies have shown that popular post-hoc techniques often underperform in medical imaging contexts compared to general computer vision tasks \cite{lime_unstable} \cite{related_gradcam}. For example, image-based methods like LIME \cite{LIME} and Grad-CAM \cite{gradcam} frequently miss clinically significant features \cite{posthocbadmedicine} \cite{breast_paper}, while recent large multimodal models such as LLaVA-Med \cite{llavamed} struggle to capture pathology-specific signals that are readily recognized by human experts \cite{textualfail}. This may stem from several factors—including limited dataset quality and noisy annotations \cite{wronglabels}—but we hypothesize that a critical and underappreciated limitation is the lack of expert involvement in the design of XAI techniques themselves.

Concept Bottleneck Models (CBMs) \cite{concept_bottleneck}, a recently popular class of ante-hoc XAI, attempt to address this by embedding human-interpretable concepts directly into the model architecture. CBMs divide prediction into two stages. First, concepts—either pre-defined or learned unsupervised—are predicted from the input image. These concepts are then used to determine the final label. This intermediary step is valuable because it is fully transparent and can be examined and evaluated by an expert. In medical diagnostics, for example, the concepts might represent clinical features in a chest X-ray, which a trained clinician can analyze to support their diagnostic process. Our evaluation includes CBMs utilizing both supervised and unsupervised concepts. 

Lung cancer remains one of the leading causes of cancer-related mortality worldwide, and early, accurate diagnosis through chest X-rays is critical. In prior work, we proposed ClinicXAI \cite{clinicxai}, a CBM-based approach for lung cancer detection using expert-defined clinical concepts to improve interpretability. While binary classification of lung cancer presence is a natural first step, real-world diagnostic workflows require distinguishing lung cancer from a range of other thoracic pathologies with overlapping imaging characteristics. To reflect this clinical reality, in this work we extend that method and introduce XpertXAI, a generalizable, expert-driven concept bottleneck model designed to scale beyond a single disease. XpertXAI is trained to detect and explain lung cancer, our clinical focus, within a multi-pathology training setup that mirrors real-world diagnostic settings, allowing us to test its broader clinical utility.

To evaluate XpertXAI, we use the large-scale public dataset MIMIC-CXR \cite{mimic-jpg} \cite{dataset} \cite{dataset_paper} of chest X-rays and associated radiological reports, comparing its performance and explanations against both post-hoc and ante-hoc XAI methods in the multiclass domain. Post-hoc techniques are applied to a 42-layer InceptionV3 model \cite{inception_arch} trained for lung pathology detection, chosen for its strong performance and widespread use in medical AI \cite{inception1} \cite{inception2} \cite{inception3} \cite{inception4}. The post-hoc methods evaluated include LIME \cite{LIME}, SHAP \cite{SHAP}, Grad-CAM \cite{gradcam}, and the textual LLM-based method CXR-LLaVA \cite{cxrllava}. For ante-hoc comparison, we include XCBs \cite{ivan}, a recent unsupervised CBM approach that infers intermediate concepts without expert input.

Our evaluation reveals that post-hoc approaches fail to capture clinically relevant information in chest X-rays, both w.r.t the medical ground truth in our dataset and the opinions of an expert radiologist. These techniques also show consistent disagreement with each other. CXR-LLaVA is notably sensitive to the wording of the input question and exhibits a high rate of false negatives on our dataset. While XCBs perform relatively well by often highlighting pathology-specific clinical concepts, they also identify less relevant concepts that do not contribute to the explanation and have a high rate of false positives according to our expert radiologist.

To test our hypothesis that expert input improves XAI quality, we designed XpertXAI using clinical concepts curated in direct collaboration with radiologists. These concepts are automatically extracted from radiology reports using standard NLP techniques and serve as supervised bottlenecks in model training. XpertXAI achieves higher classification performance than both InceptionV3 and XCBs, while also producing explanations that capture a significantly greater proportion of clinically meaningful features. According to radiologist evaluations, performed on healthy and cancerous chest X-rays in accordance with our original clinical focus, XpertXAI explanations are more accurate, reliable and aligned with expert reasoning than those produced by baseline methods.

Although XpertXAI builds directly on our prior work in lung cancer detection, its extension to multiple lung pathologies demonstrates how human-centric design principles can scale to broader diagnostic tasks. Rather than introducing a radically new architecture, XpertXAI illustrates how expert-informed concept selection and supervision can meaningfully improve both the performance and interpretability of AI models in medicine — highlighting the importance of aligning model reasoning with clinical understanding.

\section{Related Work}

\subsection{Post-Hoc Image XAI}
Post-hoc image XAI is split into perturbation- and gradient-based techniques. Perturbation-based techniques (eg. LIME \cite{LIME}, SHAP \cite{SHAP}) work by generating random masks of an image to determine which regions have the most effect on a model’s classification decision. Gradient-based techniques (eg. Grad-CAM \cite{gradcam}, LRP \cite{lrp}) instead use a forward or backward pass of a DL model to calculate the importance of each input neuron. Many recent studies apply post-hoc image XAI techniques to DL models in the domain of lung pathology classification, uncovering the lack of stability in the explanations produced \cite{lime_unstable} \cite{counterfactual}. Explanations from these techniques have been shown to disagree with each other, with the medical ground truth of the dataset and with the opinions of a radiologist \cite{breast_paper} \cite{collab_paper} \cite{related_gradcam} \cite{covid_cxnet} \cite{covid}. This has raised concerns regarding their use in the diagnostic domain \cite{posthocbadmedicine2}. Post-hoc explanations routinely highlight irrelevant regions of medical scans while missing clinical features, despite the high classification performances of the models they are applied to \cite{posthocbadmedicine}. 

\subsection{Post-Hoc LLM-Based XAI}
The recent popularity of ChatGPT \cite{chatgpt} and associated open-source large language models (LLMs) has given rise to textual XAI techniques in the form of multimodal models like LLaVA \cite{llava}, which connect vision encoders and LLMs, usually in a chatbot format. These techniques use Visual Question Answering (VQA) - the process of building models which can answer questions about input images \cite{vqa1} \cite{vqa2}.  In the medical domain, biomedical chatbots such as LLaVA-Med \cite{llavamed} and CXR-LLaVA \cite{cxrllava} have been introduced. In this work we examine the effectiveness of CXR-LLaVA, which is pre-trained on multiple chest X-ray datasets including the one used in this work, for generating clinically relevant explanations for Chest X-Rays.

\subsection{Concept Bottleneck Models} \cite{concept_bottleneck} introduce CBMs, which essentially split the traditional classification pipeline into 2 separate models. The first model takes an image and predicts the presence of a list of pre-determined concepts, and the second takes these concepts and predicts the output label. These concepts act as a convenient middle step to the pipeline, giving the user valuable insight into the decision making process. A major drawback of CBMs is the need for a fully annotated training dataset, which for domains such as medical imaging is infeasible and expensive. There are many approaches for handling this problem in literature. \cite{posthoccbm} apply CBMs in a post-hoc format, which can be applied to any classifier. Other approaches such as label-free CBMs \cite{lcbms} and language model guided CBMs \cite{labo} leverage LLMs such as GPT-3 \cite{gpt3} to produce domain-specific concepts, and the CLIP model \cite{clip} to automatically label the images with these concepts. \cite{visualcbm} highlight the fact that these LLM-based approaches tend to generate non-visual concepts, and include visual activations to attempt to rectify this. \cite{ivan} introduce a cross-modal CBM (XCBs) which instead extracts concepts from radiology reports associated with chest X-rays using an unsupervised cross-attention mechanism. 

We note that these approaches generate concepts in an unsupervised manner, without expert input, and their clinical usability in specific medical domains is unclear. In order to test our hypothesis that expert input has a positive impact on the clinical relevance of explanations, we introduce our own expert-driven CBM approach in this work. Instead of extracting unsupervised concepts, we leverage common NLP techniques to automatically extract key phrases indicative of pathologies from the radiology reports associated with the chest X-rays in our dataset. The phrases we extract were decided under direct radiologist guidance, meaning that our concepts have direct diagnostic interpretability, and are fully customisable by the expert user. 

\section{Evaluation of Existing XAI}

%In this Section, we describe the dataset and the existing XAI techniques used in our evaluation. We begin by evaluating the clinical relevance of explanations generated by several existing XAI techniques in the context of lung cancer detection in chest X-rays.

\subsection{Dataset}\label{dataset_Section}

We use the MIMIC-CXR dataset from PhysioNet, which comprises chest X-rays and associated free-text radiology reports \cite{dataset} \cite{dataset_paper} \cite{physionet}. Although the dataset includes X-rays from multiple views (PA, AP, and lateral), we focus exclusively on the Posterior-Anterior (PA) view to minimize confounding factors \cite{pa_usage2}. Under radiologist guidance, we select six clinically significant labels from the original fourteen: Healthy (No Finding), Lung Cancer (Lung Lesion), Pneumonia, Pneumothorax, Pleural Effusion, and Cardiomegaly. These conditions were chosen for their relevance to real-world diagnostic practice and their representation of general pathological categories rather than symptom-specific labels such as Consolidation or Lung Opacity. Due to a substantial class imbalance—most cases are labelled as Healthy—we apply the One-Sided Selection undersampling technique \cite{onesided} to mitigate its impact on classification performance. The final dataset comprises 35,893 image-report pairs, which we split into training, validation, and testing sets using an 80/10/10 ratio. All images are resized to 512×512 pixels, normalized, and cropped to remove black borders.

\subsection{XAI Techniques}

For our evaluation, we select XAI techniques from various families. For post-hoc image XAI, we use gradient-based Grad-CAM \cite{gradcam} and perturbation-based LIME \cite{LIME} and SHAP \cite{SHAP}, which are well-documented and widely used. For textual LLM-based XAI, we employ CXR-LLaVA \cite{cxrllava}, a high-performing LLaVA-based method specialized in chest X-ray pathologies. For ante-hoc XAI, we use XCBs \cite{ivan}, a recent CBM-based method that learns unsupervised concepts from chest X-rays and radiology reports, demonstrating promising results compared to other CBMs \cite{lcbms}.
LIME, SHAP and Grad-CAM are implemented using their respective Python libraries and applied post-hoc to an InceptionV3 model \cite{inception_arch}, which was trained on our MIMIC-CXR dataset, a widely-used dataset for chest X-ray pathology detection \cite{inception1} \cite{inception3}. Implementations of CXR-LLaVA and XCBs are publicly available. All experiments were performed with an NVIDIA GTX 1060 6GB GPU.

\begin{figure}[tb]
\centering
\begin{subfigure}{0.45\textwidth}
    \includegraphics[width=\textwidth]{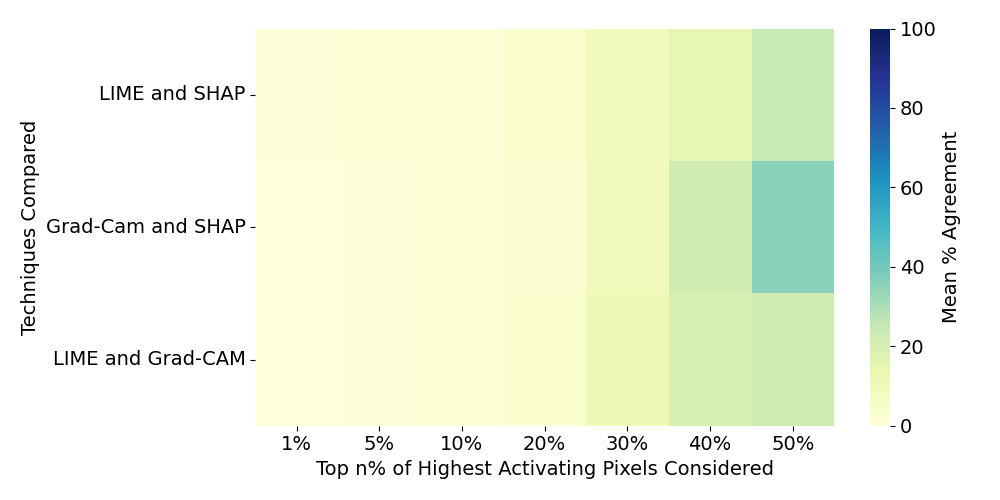}
\end{subfigure}
\begin{subfigure}{0.45\textwidth}
    \includegraphics[width=\textwidth]{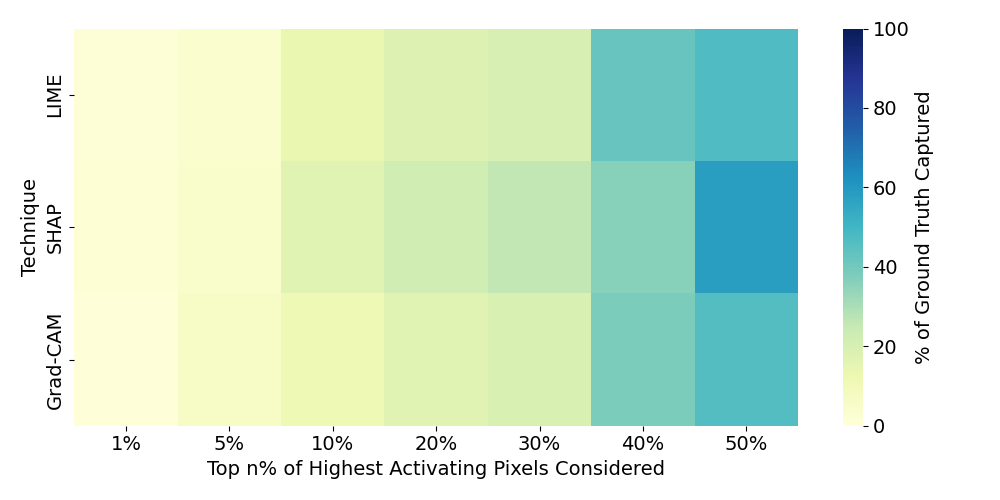}
\end{subfigure}
\caption{For LIME, SHAP and Grad-CAM, (a) shows mean pixel overlap between techniques on the MIMIC-CXR test set. (b) shows mean medical ground truth captured on the VinDr-CXR test set.}
\label{fig:pixeloverlap}
\end{figure}

\subsection{Observations}

\subsubsection{Image-Based XAI}\label{sec:radio}

For LIME, SHAP and Grad-CAM, which generate explanations based solely on input images, we evaluate: 1) the agreement between techniques (multiclass), 2) the clinical relevance of explanations based on overlap with ground truth annotations (multiclass), and 3) the alignment with expert radiologist opinions (lung cancer only). While our technical evaluation spans multiple thoracic diseases, our clinical validation remains focused on lung cancer, aligning with our central clinical focus. These approaches are applied post hoc to a standard InceptionV3 model, which achieves a high macro-average F1 score of 0.74 across the six classes in our dataset.

To assess agreement between techniques, we analyze the similarity of the top n\% highest-activating pixels across the MIMIC-CXR test set for various values of n. These pixels represent the most salient regions influencing the model's classification decisions. If the explanations reliably highlight meaningful image features, we would expect a high degree of overlap between methods for small n. As shown in Figure~\ref{fig:pixeloverlap}(a), however, agreement remains consistently low across LIME, SHAP, and Grad-CAM, even in this multiclass diagnostic setting. This suggests that these techniques may not be reliably isolating shared diagnostic cues across pathologies.

\begin{figure*}[tb]
\centering
\begin{subfigure}{0.45\textwidth}
    \includegraphics[width=\textwidth]{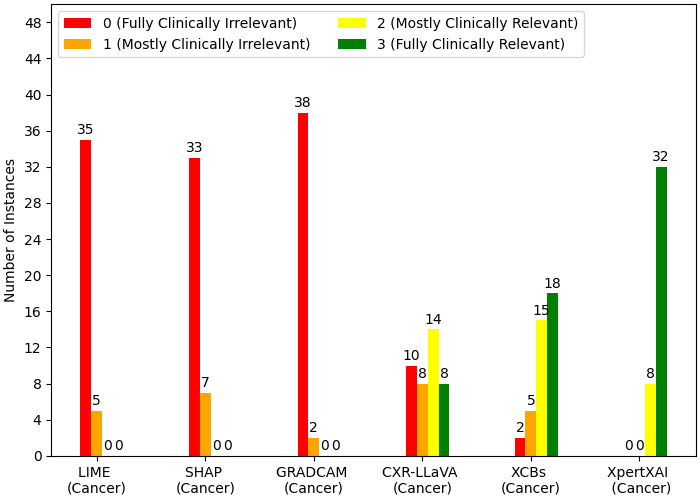}
\end{subfigure}
\begin{subfigure}{0.45\textwidth}
    \includegraphics[width=\textwidth]{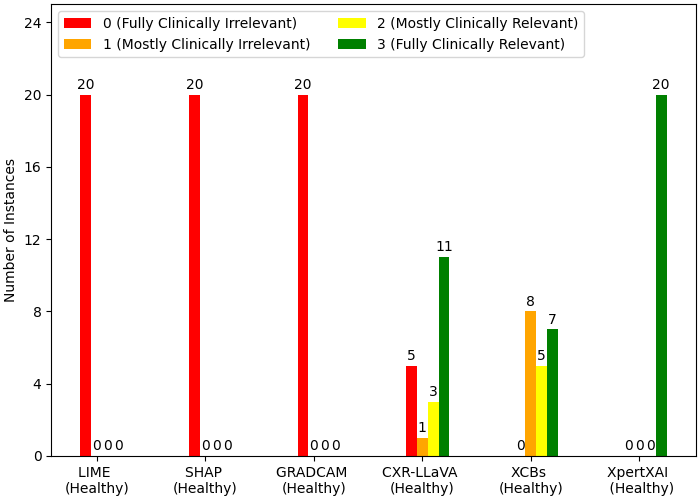}
\end{subfigure}
    \caption{Analysis by an expert radiologist of explanations generated for a subset of 40 cancerous and 20 healthy chest X-rays. The expert was asked to score explanations between 0 and 3 (see legend). XpertXAI refers to the DT architecture.}
\label{fig:expert}
\end{figure*}

To evaluate clinical relevance, we compare explanations to ground truth pathology locations using a subset of the VinDr-CXR dataset \cite{vindata}, which includes radiologist-annotated bounding boxes for multiple thoracic conditions. Since our goal is to assess technical explanation quality across a range of diseases, we extend this analysis to include all available annotated pathologies that correspond to our six target pathology labels – all except Pneumonia. We compute the proportion of bounding box pixels captured within the top n\% of the explanation maps across this evaluation set of 921 chest X-rays. As shown in Figure~\ref{fig:pixeloverlap}(b), LIME, SHAP and Grad-CAM often fail to localize clinically relevant regions.

\begin{figure*}[tb]
\centering
\includegraphics[width=0.7\textwidth]{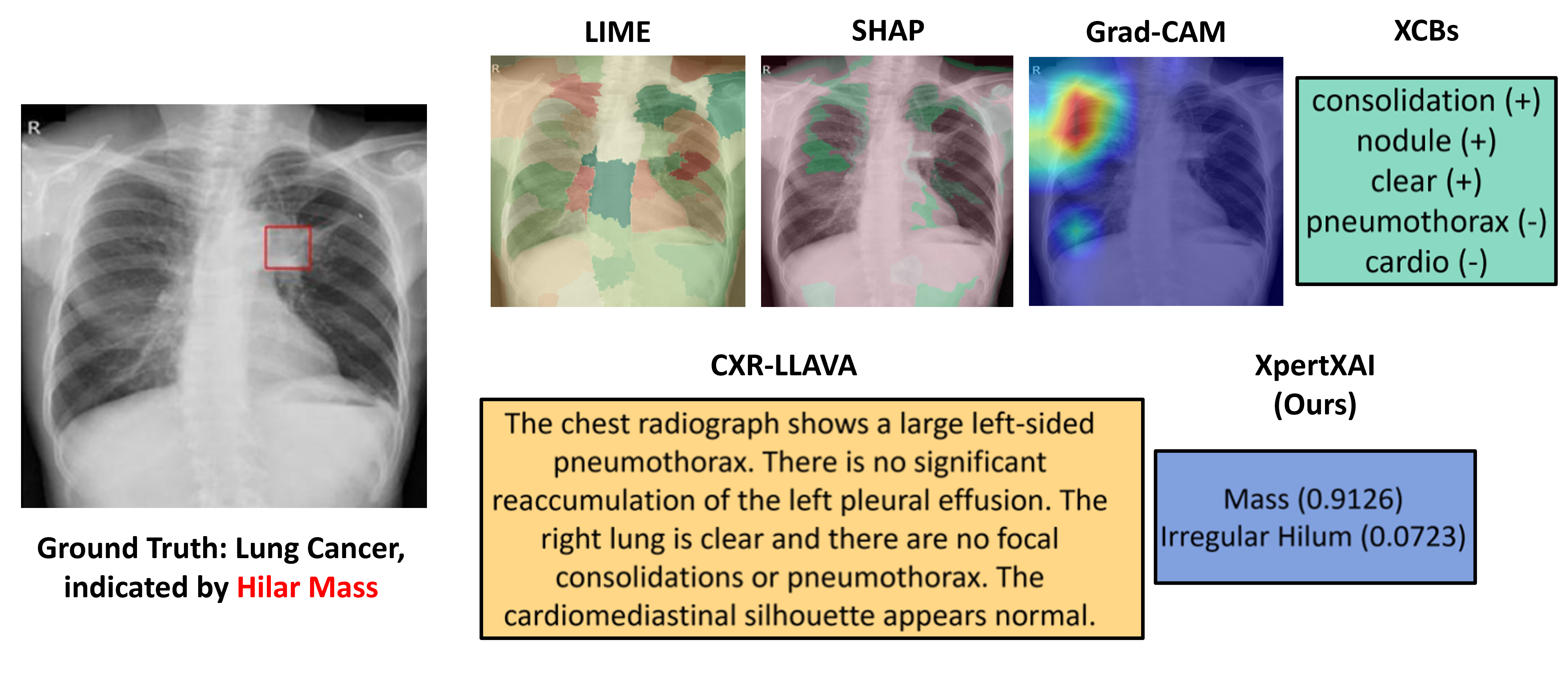}
\caption{Explanations generated by each XAI technique for a cancerous chest X-ray. (a) shows the ground truth hilar mass. (b) shows LIME (most important = intense green). (c) shows SHAP (most important = green). (d) shows Grad-CAM (most important = red). (e) shows XCBs (concepts with the 5 highest absolute values, positive (+) or negative (-)). (f) shows the radiology report generated by CXR-LLaVA. (g) shows XpertXAI (DT architecture), with the top 2 scoring concepts.}
\label{fig:bigexample}
\end{figure*}

\paragraph{Comparisons to Expert Opinions (Lung Cancer Only)}
To assess alignment with human clinical reasoning, we compare explanations from LIME, SHAP, and Grad-CAM to the opinions of an expert radiologist, focusing exclusively on lung cancer cases. A radiologist with over 10 years of experience reviewed explanations for 60 chest X-rays — 40 cancerous and 20 healthy. This dataset is small due to radiologist availability, and will be expanded in future works. Each image was evaluated across all six XAI methods used in this study. The expert was asked to score each explanation between 0 and 3, where 0 indicates the explanation is fully clinically irrelevant, 1 indicates it is mostly clinically irrelevant, 2 indicates it is mostly clinically relevant, and 3 indicates that it is fully clinically relevant.
As shown in Figure~\ref{fig:expert}, post-hoc explanations for both cancerous and healthy images predominantly scored 0 across all techniques. The radiologist noted that image-based techniques frequently highlighted anatomically implausible regions such as clear lung fields or areas outside the lungs. This suggests a limited degree of clinical utility in their current form. An example of such mislocalization is shown in Figure~\ref{fig:bigexample}(b–d), where LIME, SHAP and Grad-CAM scored 0, 1, and 0 respectively for a clearly cancerous scan. This reinforces concerns about the clinical utility of current XAI methods, particularly in critical tasks like lung cancer detection.

\subsubsection{LLM-Based XAI}\label{cxr-lava-sec}

CXR-LLaVA is a vision-language model designed to generate radiology-style reports and answer diagnostic questions based on chest X-ray input. In this study, we evaluate CXR-LLaVA in three key areas: 1) response consistency across semantically equivalent questions, 2) clinical relevance of generated reports based on proportion of ground truth clinical concepts captured (multiclass), and 3) alignment with expert radiologist judgment in the context of lung cancer detection.

To assess response consistency, we pose multiple semantically similar prompts for chest X-rays in our evaluation set. These include neutral queries (e.g., Do any clinical features of this chest X-ray indicate lung cancer?) and diagnosis-assuming prompts (e.g., Which clinical features of this chest X-ray indicate lung cancer?). We find that while neutral prompts typically result in appropriate "no finding" responses on healthy images, diagnosis-assuming prompts often elicit confident but incorrect pathology-related descriptions—despite the absence of pathology. This suggests that CXR-LLaVA may be vulnerable to prompt-induced hallucinations when prompt phrasing implies a disease presence.

To evaluate clinical relevance, we followed a method similar to \cite{cxrllava}, applying the same NLP concept extraction techniques to both ground truth MIMIC-CXR reports and CXR-LLaVA-generated reports. We then compared the proportion of shared clinical concepts. The clinical concepts, selected by a consultant radiologist, represent key phrases used by experts in describing healthy chest X-rays, or X-rays indicative of our five target pathologies (Lung Cancer, Cardiomegaly, Pneumonia, Pneumothorax, Pleural Effusion). We found that CXR-LLaVA’s reports frequently produced false negatives specifically for cancer diagnosis, instead emphasizing pneumothorax or cardiomegaly, likely due to imbalances in the training dataset \cite{cxrllava}. On our MIMIC-CXR test set of 3590 chest X-rays and radiology reports, CXR-LLaVA achieved an overall concept F1 score of \textbf{0.658}.

\begin{table}[tb]
%\small
\centering
\begin{tabular}{lrrr}
Model & \multicolumn{3}{c}{\textbf{Accuracy}} \\
%\cline{2-7}
& Precision & Recall & F1 Score \\
%& & & & & & &\\
\midrule
XpertXAI & \textbf{0.772} & \textbf{0.882} & \textbf{0.823} \\
CXR-LLaVA & 0.713 & 0.418 & 0.527 \\
XCBs & 0.598 & 0.854 & 0.703\\
\bottomrule
\end{tabular}
\caption{
Proportions of ground truth clinical concepts captured by CXR-LLaVA, XCBs, and our approach XpertXAI (DT architecture), evaluated on the 3590 image-report pair MIMIC-CXR multiclass test set.}
\label{textual_comps}
\end{table}

We assess alignment with expert opinion through a radiologist-led lung cancer-specific evaluation of generated reports for 60 chest X-rays (40 cancerous, 20 healthy), as with post-hoc image XAI. As shown in Figure~\ref{fig:expert}, CXR-LLaVA performs fairly well on healthy cases but often receives low scores on cancerous scans. The expert highlighted common failures such as omission of key cancer findings (e.g., small nodules, hilar masses), introduction of unrelated conditions (e.g., scoliosis, pneumothorax), and logical inconsistencies within reports. An example is shown in Figure~\ref{fig:bigexample}(f), where a scan with a visible hilar mass is misreported as clear, and a pneumothorax is incorrectly suggested, resulting in a clinical relevance score of 0.

\subsubsection{CBM-Based XAI}

XCBs adopt a CBM architecture utilizing an unsupervised learned concept space, rather than introducing direct expert input to concept design. We evaluate the clinical utility of concept-based explanations produced by XCBs in two ways: 1) clinical relevance based on proportion of ground truth clinical concepts captured (multiclass), and 2) agreement with expert radiologist judgment on cancer-related cases.

To assess clinical feature coverage, we apply the same NLP-based concept extraction strategy as in our CXR-LLaVA analysis. Clinical concepts are extracted from ground truth MIMIC-CXR reports and compared to the five highest absolute-scoring concepts predicted by XCBs for each chest X-ray, following \cite{ivan}. XCBs outperformed CXR-LLaVA, achieving an overall concept F1 score of \textbf{0.799}, however the technique has a significantly lower Precision score, indicating a high rate of false positives.

For expert assessment, a consultant radiologist evaluated XCBs explanations for 60 chest X-rays (40 cancerous, 20 healthy), as with previous approaches. As shown in Figure~\ref{fig:expert}, XCBs perform well on cancerous scans, frequently achieving scores of 2 or 3. The expert noted that XCBs were particularly effective at identifying salient abnormalities such as masses, but occasionally included irrelevant or misleading concepts (e.g., false mentions of consolidation). For example, Figure~\ref{fig:bigexample}(e) shows an explanation which was scored 2 by the expert, in which a correctly identified hilar mass was accompanied by an erroneous reference to consolidation. Performance on healthy images was less reliable, with the model frequently surfacing spurious features, leading to lower expert scores and a higher false positive rate. This suggests that while XCBs offer a promising pathway toward interpretable AI, additional refinement is needed to improve specificity on normal scans.

\begin{figure}[tb]
\centering
    \includegraphics[width=0.49\textwidth]{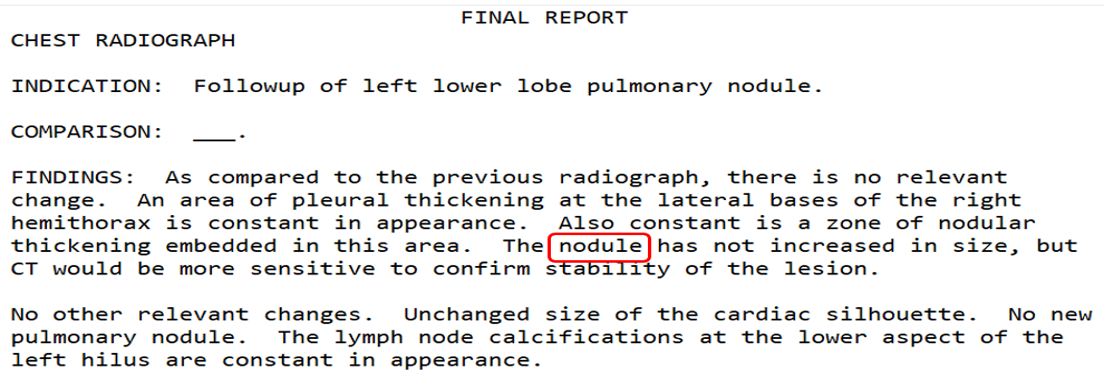}
\caption{Example of a cancerous radiology report from the MIMIC-CXR dataset. Clinical concepts extracted (`Nodule') are highlighted by bounding box. Note the negative mention in the final paragraph is not extracted.}
\label{fig:report}
\end{figure}

\section{XpertXAI: An Expert-Driven Concept Bottleneck Model}

\begin{figure*}[tb]
\centering
    \includegraphics[width=0.7\textwidth]{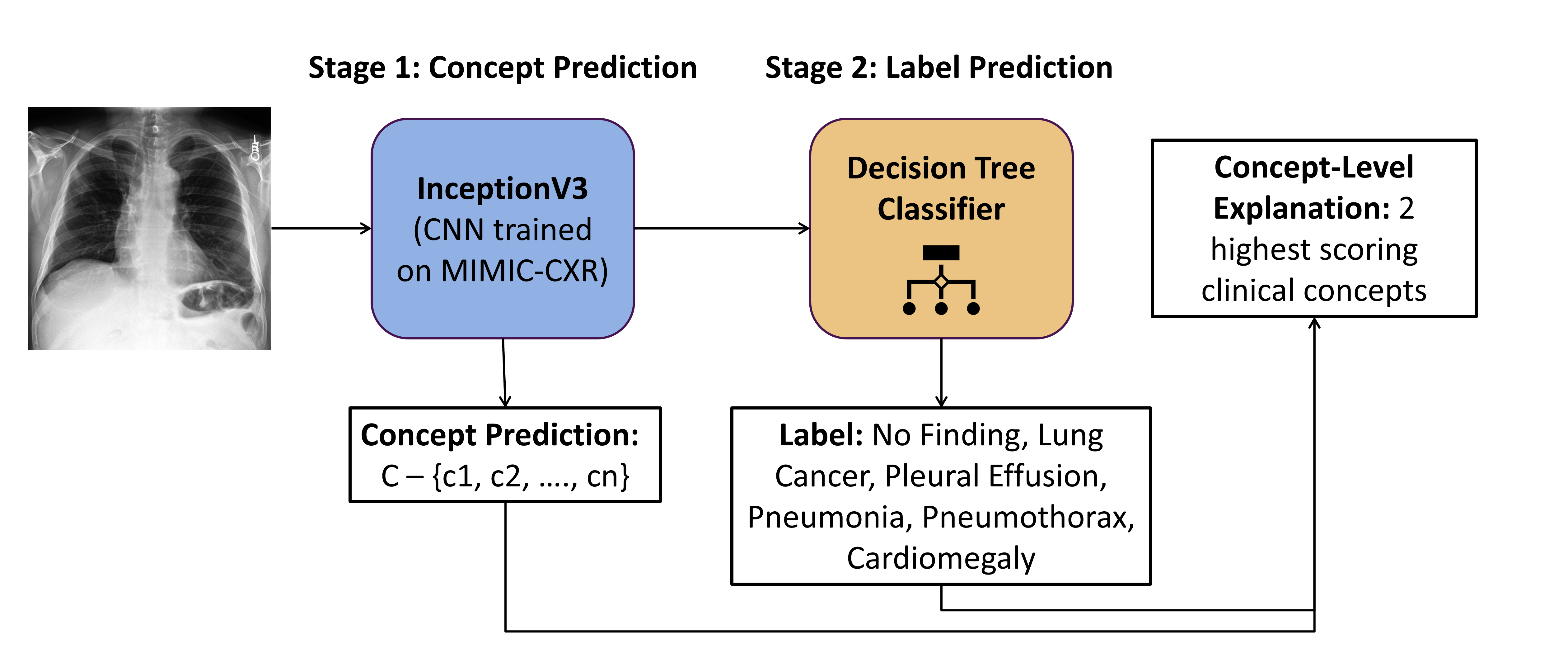}
    \caption{The pipeline for XpertXAI: We take a chest X-ray as input, which is fed into a trained concept prediction model, producing prediction scores for a pre-set list of clinical concepts. These scores are then input to a trained label prediction model, which outputs the binary classification label. Explanations supporting model decisions are shown as the two highest scoring concepts.}
    \label{fig:inference}
\end{figure*}

Existing XAI techniques often perform poorly in pathology detection from chest X-rays, failing to capture clinically relevant features. To explore the potential benefits of expert input on XAI performance and explanation quality, we introduce a CBM-based approach that uses clinical concepts selected by an expert radiologist. Unlike XCBs, which identify concepts unsupervised, XpertXAI predicts the presence of these expert-chosen concepts. This section describes XpertXAI and compares its classification performance and clinical relevance of explanations with those of CXR-LLaVA and XCBs.

\subsection{Clinical Concept Extraction}

Each chest X-ray in the MIMIC-CXR dataset is linked to a free-text radiology report (Figure \ref{fig:report}), similar to clinical practice. Given the variability in report structures, data cleaning was performed under radiologist guidance, focusing on relevant sections (FINDINGS and IMPRESSION) \cite{mimic-jpg}. We removed punctuation and extraneous words to generate a list of cleaned sentences for analysis. 
Using a list of radiologist-provided clinical concepts in the form of key phrases, we identified which were present in the cleaned reports, excluding negative mentions. These phrases were grouped by meaning into clinical concepts by our expert radiologist to reduce sparsity and redundancy. This information was then used to automatically convert reports into binary values indicating each concept's presence. This domain-specific extraction method outperforms automated techniques like CheXpert \cite{chexpert} by reducing false positives through the consideration of negative mentions.

Our final concept list contains 1 Healthy concept (Unremarkable), 5 Lung Cancer concepts (Mass, Nodule, Irregular Hilum, Adenopathy, Irregular Parenchyma), 4 Pneumonia concepts (Pneumonitis, Consolidation, Infection, Opacities), 4 Pleural Effusion concepts (Effusion, Fluid, Costophrenic Angle, Meniscus Sign), 1 Cardiomegaly concept (Enlarged Heart) and 2 Pneumothorax concepts (Absent Lung Markings, Irregular Diaphragm). These concepts, and their associated report phrases, are provided on Github \footnote{https://github.com/AmyRaff/concept-explanations}. A consultant radiologist analyzed a random sample of reports belonging to each target label, highlighting key phrases relevant to expert diagnosis. Single expert labeling, a common practice in radiology AI research \cite{hosny2018artificial}, was used. Since our clinical concepts are derived from pre-annotated reports, potential bias is minimized.
The list of clinical concepts associated with a pathology is well understood in the medical community and is straightforward for an expert to provide. The consultant radiologist in our study needed approximately 20 minutes to provide the list of concepts for our six target labels.
It is worth noting that once the list of clinical concepts for a pathology has been provided, expert input is no longer needed – we automatically annotate our dataset with the presence of these concepts within the report and use them to train our model. Cost associated with obtaining expert input for XpertXAI is low as it is just the list of terms/concepts associated with a given pathology, which can be reused. 

\subsection{Architecture}

As in the original CBM work \cite{concept_bottleneck}, we split the traditional classification pipeline into two separate models, shown in Figure~\ref{fig:inference}. We use the Independent CBM architecture as we are focusing on the clinical usability of explanations generated using the intermediate concept step, rather than the joint CBM architecture which merges the process into one end-to-end model.
The first is the concept prediction model, which during inference takes a chest X-ray, and outputs a prediction score for each of a pre-determined list of concepts. This model is trained using both chest X-ray images and associated radiology reports from the MIMIC-CXR training set. As in the original CBM work we use an InceptionV3 architecture, trained with a batch size of 64 and learning rate of 0.001.
The second model is the label prediction model, which during inference uses the concept prediction scores generated by the previous model to predict the image label. This model is trained on the radiology reports and pathology labels from our training dataset. The original CBM work uses a Multilayer Perceptron (MLP) model architecture for label prediction. We also experiment with Support Vector Machines (SVMs) and Decision Trees (DTs) to understand which architecture is most suitable for Chest X-rays. We find DTs have the best performance. 

\section{Results}

\begin{figure}[tb]
\centering
    \includegraphics[width=0.38\textwidth]{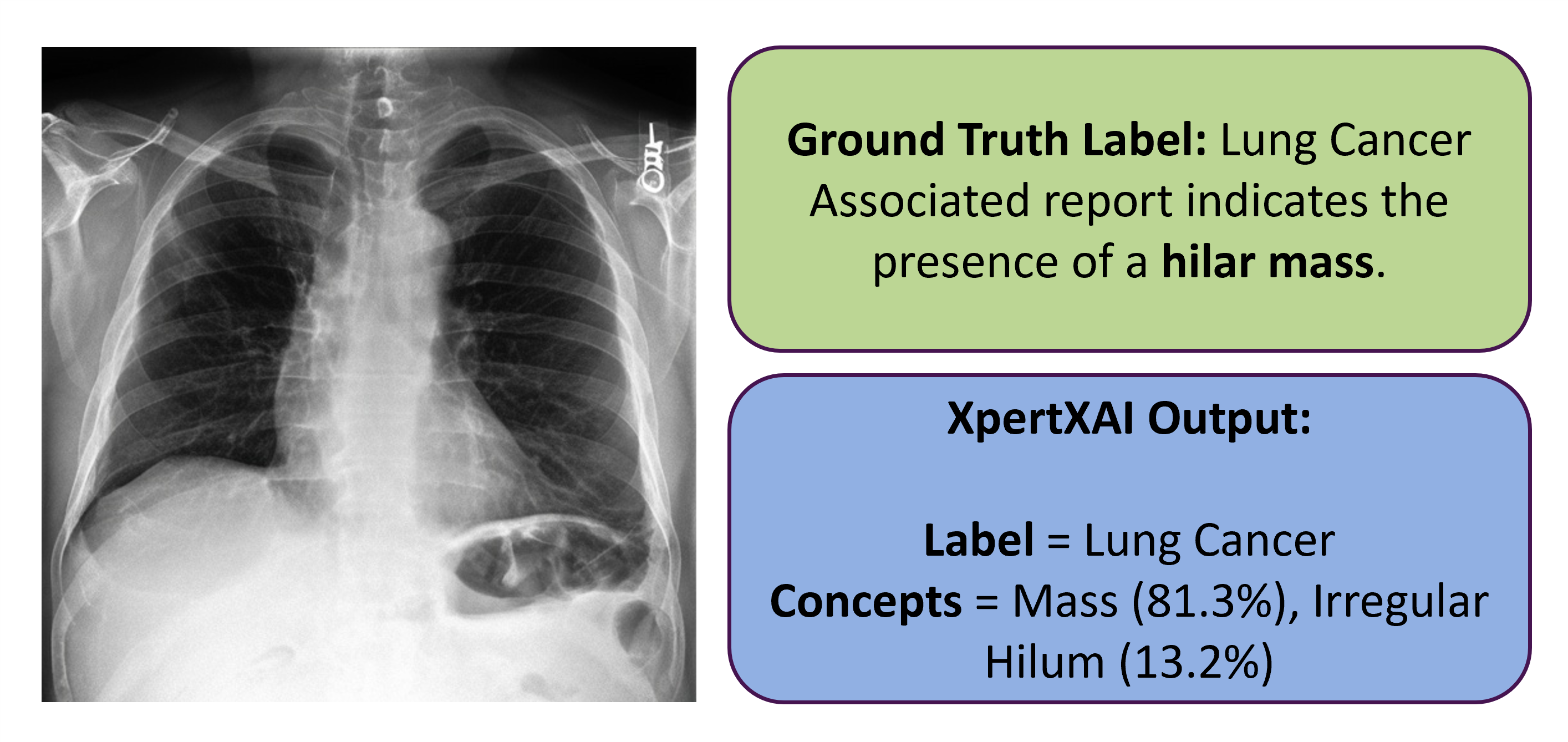}
\caption{Example correct XpertXAI explanation generated on a cancerous X-ray. The 2 highest scoring concepts are shown as an explanation.}
\label{fig:exps}
\end{figure}

\subsection{Concept Prediction Performance}

Since XpertXAI generates text-based explanations in the form of the two highest-confidence clinical concepts for a given chest X-ray, we evaluate its interpretability using the same methodology applied to other text-based approaches in this study, CXR-LLaVA and XCBs. Specifically, we assess its ability to reliably capture ground truth clinical concepts in its explanations. The results, presented in Table \ref{textual_comps}, show that XpertXAI has an F1 score of \textbf{0.842} for ground truth concept prediction on the MIMIC-CXR test set of 3590 chest X-rays and associated reports. A successful prediction is defined as the ground truth concept being one of the 2 highest scoring concepts in the prediction, as this is what we use as an explanation. This performance surpasses that of XCBs (0.799) and CXR-LLaVA (0.658). These findings support our hypothesis that incorporating expert input when designing an explainable model framework enhances its interpretability. By ensuring that the concept space is clinically relevant, XpertXAI aligns more closely with real-world clinical data, thereby increasing its interpretability and therefore usability to medical professionals.
Our earlier model, ClinicXAI, trained exclusively for binary lung cancer classification, achieved 100\% concept prediction accuracy on a smaller dataset. While expanding to multiclass training slightly reduces concept prediction performance, likely due to the larger concept space, XpertXAI maintains superior interpretability and clinical utility when compared to other approaches.

\subsection{Label Prediction Performance}

Table \ref{label_comps} summarizes the label prediction performance of XpertXAI using three label classifier architectures—MLP, Decision Trees (DT), and SVM — compared against the state-of-the-art CBM-based XAI approach, XCBs, and the standard InceptionV3 model. The InceptionV3 model, previously used to benchmark post-hoc image-based XAI in this study, serves as a baseline. All models were trained on our curated multiclass MIMIC-CXR dataset. Among the XpertXAI variants, the DT-based architecture achieved the highest performance, achieving an F1 score of \textbf{0.866} on the MIMIC-CXR test set comprising 3,590 chest X-rays. This outperforms the InceptionV3 baseline and the XCB model, highlighting XpertXAI’s superior capability in clinically grounded label prediction over general-purpose and unsupervised concept-based alternatives.
Our earlier model, ClinicXAI, trained for binary lung cancer classification, achieved a pathology prediction F1 score of 0.891. Expanding to multiclass training has had a negligible impact on label prediction performance, supporting the scalability of expert-centric explainable models to additional pathologies.

\begin{table}[tb]
%\small
\centering
\begin{tabular}{lrrr}
%\cline{2-7}
\toprule
Model & Precision & Recall & F1 \\
%& & & & & & &\\
\midrule
Standard (InceptionV3) & 0.555 & 0.731 & 0.631 \\
XCBs & 0.821 & 0.766 & 0.792 \\
XpertXAI with DT & \textbf{0.832} & \textbf{0.892} & \textbf{0.861} \\
XpertXAI with SVM & 0.656 & 0.843 & 0.738 \\
XpertXAI with MLP & 0.722 & 0.639 & 0.678 \\
\bottomrule
\end{tabular}
\caption{Label classification performance of the standard InceptionV3 model, XCBs model, and XpertXAI with multiple label classifier architectures, on the MIMIC-CXR test set.}
\label{label_comps}
\end{table}

\subsection{Alignment with Expert Opinions}

While all methods evaluated in this work were trained in a multiclass setting, we exclusively evaluate alignment with expert radiologist opinions for the lung cancer class to maintain fidelity with our research goals. XpertXAI demonstrates superior performance compared to all other evaluated approaches in our expert radiologist assessment of 60 chest X-rays. Explanations for all healthy scans and the majority of cancerous scans are deemed fully clinically relevant by the expert. The only critique was the occasional mislabeling of masses as nodules and vice versa.
Figure \ref{fig:bigexample} illustrates XpertXAI outperforming all other methods evaluated (LIME, SHAP, Grad-CAM, CXR-LLaVA, and XCBs). It shows explanations for a chest X-ray with a hilar mass, where XpertXAI correctly identifies "Mass" as the highest scoring concept and "Irregular Hilum" as the second, strongly indicating the presence of a hilar mass.

\section{Conclusion}

We evaluated the clinical relevance of leading explainable AI methods for chest X-ray pathology detection, using both post-hoc and ante-hoc approaches. While the technical evaluation covered six lung pathologies, expert validation focused specifically on lung cancer—the central diagnostic task of this work. Existing XAI methods frequently failed to produce clinically meaningful explanations, missing key diagnostic features and diverging from expert reasoning. To address this, we introduce XpertXAI, an expert-guided concept bottleneck model that generates interpretable predictions grounded in clinically relevant concepts. XpertXAI outperformed post-hoc and unsupervised CBMs in both explanation quality (measured by proportion of ground truth clinical concepts captured) and classification performance. Crucially, it produced the most clinically relevant explanations for lung cancer cases, as assessed by a consultant radiologist. These findings underscore the importance of incorporating domain expertise into model design for trustworthy AI in high-stakes settings like lung cancer detection.
We acknowledge that our current focus is limited to chest X-rays, and we plan to adapt our work to other imaging modalities, such as CT and MRI. Our expert evaluation is also limited, though important, and we will expand upon this in future work. While we employ an InceptionV3 model, which might raise concerns about the generalizability of our findings, this model has consistently demonstrated strong performance in the field and serves as the backbone of our CBM architecture.

%% The file named.bst is a bibliography style file for BibTeX 0.99c
\bibliographystyle{named}
\bibliography{ijcai25}

\end{document}